\documentclass[letterpaper, 10 pt, conference]{ieeeconf}  

\IEEEoverridecommandlockouts                              

\usepackage{times}
\usepackage{graphicx}
\usepackage{amsmath}
\usepackage{mathabx}
\usepackage{algorithm}
\usepackage{algpseudocode}
\usepackage[T1]{fontenc}
\usepackage{array}
\usepackage{booktabs,ragged2e}
\usepackage{float}
\usepackage[flushleft]{threeparttable}
\usepackage{color,soul}
\usepackage{multicol}
\usepackage[bookmarks=true]{hyperref}
\usepackage{svg}

\title{\LARGE \bf
 Cosserat Rod Modeling and Validation for a Soft Continuum Robot with Self-Controllable Variable Curvature }

\author{Xinran~Wang,~\IEEEmembership{Student~Member,~IEEE} and Nicolas~Rojas,~\IEEEmembership{Member,~IEEE}
\thanks{Xinran Wang and Nicolas Rojas are with the REDS Lab, Dyson School of Design Engineering, Imperial College London, 25 Exhibition Road, London, SW7 2DB, UK
{\tt\small (xinran.wang20, n.rojas)@imperial.ac.uk}}%
}

\begin{document}

\maketitle
\thispagestyle{empty}
\pagestyle{empty}

\begin{abstract}

This paper introduces a Cosserat rod based mathematical model for modeling a self-controllable variable curvature soft continuum robot. This soft continuum robot has a hollow inner channel and was developed with the ability to perform variable curvature utilizing a growing spine. The growing spine is able to grow and retract while modifies its stiffness through milli-size particle (glass bubble) granular jamming. This soft continuum robot can then perform continuous curvature variation, unlike previous approaches whose curvature variation is discrete and depends on the number of locking mechanisms or manual configurations. The robot poses an emergent modeling problem due to the variable stiffness growing spine which is addressed in this paper. We investigate the property of growing spine stiffness and incorporate it into the Cosserat rod model by implementing a combined stiffness approach. We conduct experiments with the soft continuum robot in various configurations and compared the results with our developed mathematical model. The results show that the mathematical model based on the adapted Cosserat rod matches the experimental results with only a 3.3\% error with respect to the length of the soft continuum robot. 
 
\end{abstract}

\section{Introduction}

A continuum robot is a highly flexible and adaptable robotic system, capable of achieving complex motions even in the most constrained environments. The use of silicone rubber, fabrics or soft TPU materials \cite{guan2023trimmed}\cite{fras2015new}\cite{haggerty2023control} has made the soft continuum robot even more adaptive to its surroundings, enabling it to perform even more complex motions. Recently, there has been a surge of interest in continuum robots due to its variable curvature capability, making it an incredibly versatile and powerful tool for a wide range of applications \cite{rao2023modeling}\cite{ma2023inspired}\cite{zhang2023preprogrammable}. By changing the local stiffness of the continuum robot, the ability to achieve variable curvature can be achieved. This allows for the continuum robot to have different curvatures, even when receiving the same control inputs, resulting in a highly versatile and adaptable system.

\begin{figure}[htbp]
 \centering
 \includegraphics[width=\linewidth]{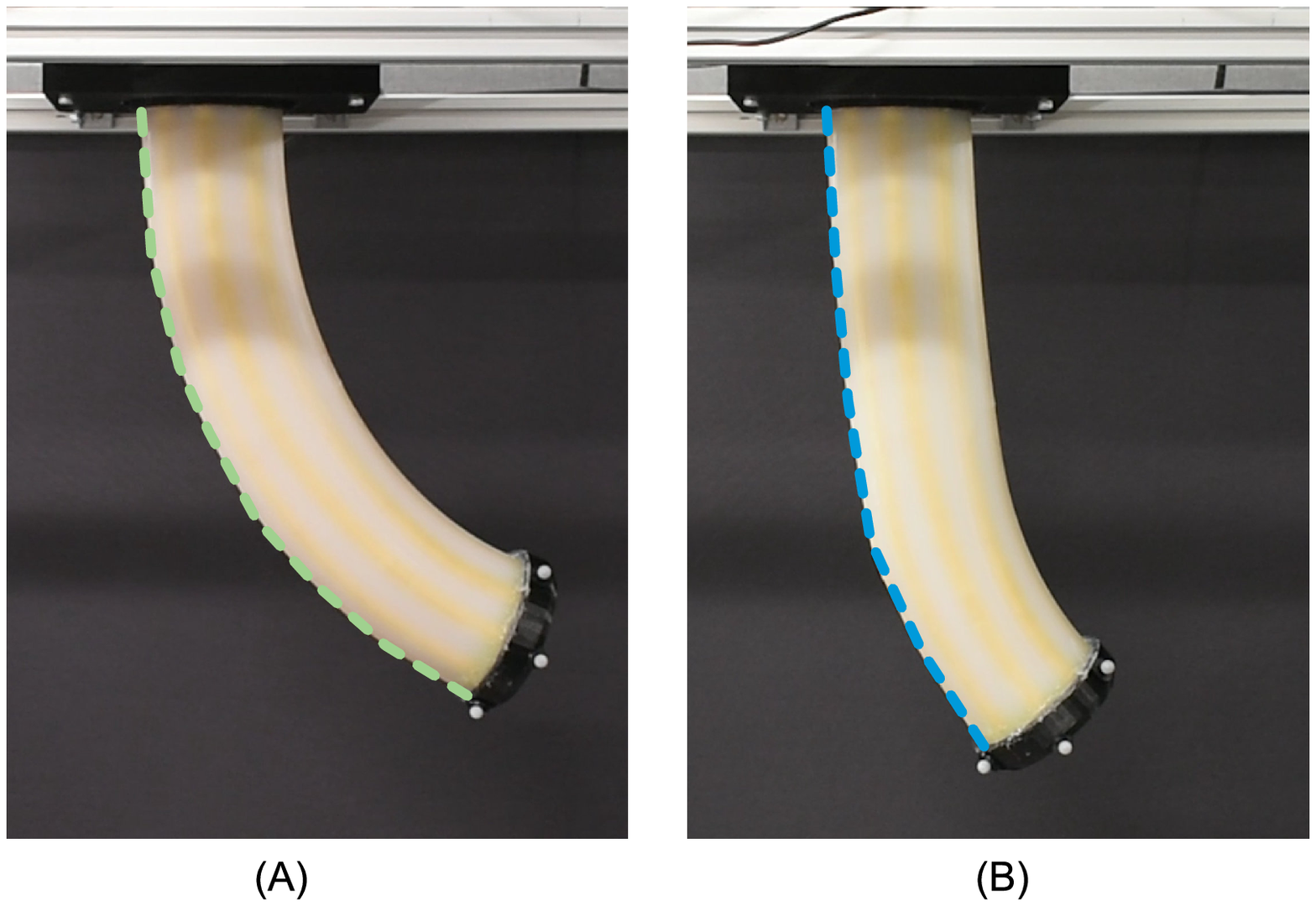} 
 \caption{Examples of variable curvature of the soft continuum robot under study, which is based on a variable stiffness self-growing spine. (A) Curvataure of the robot with 0 cm growing spine length when pressurized at 250 kPa. (B) Curvature of the robot with 30 cm growing spine length when pressurized at 250 kPa.}
 \label{fig:1}
\end{figure}

In order to achieve variable curvature in traditional continuum robot, many techniques have been applied like using shape memory alloy (SMA) \cite{yang2020geometric}, locking mechanism using magnets \cite{pogue2022multiple} or screws \cite{rao2023modeling}, pressurized fluid chambers \cite{stella2023prescribing} and pre-configuring different springs stiffness along the continuum robot \cite{zhang2023preprogrammable}. The high adaptability and relatively low stiffness of the soft continuum robot have led to the exploration of more options for altering its overall stiffness. The stiff-flop manipulator, which incorporates granular jamming in each section, is an effective technique for overall stiffness change \cite{cianchetti2014soft}. Additionally, fiber jamming and layer jamming techniques are presented as viable solutions for stiffness change \cite{arleo2023variable}\cite{clark2022malleable}\cite{kang2023soft}. In fact, the most recent development shows that inserting disks into the continuum robot can allow for stiffness change \cite{ma2023inspired}. However, it is important to note that these curvature changes are currently limited in terms of the number of sections on the continuum robot or locking mechanism used, as they are limited in a discrete fashion. In our previous work, we have developed a novel soft continuum robot with the ability to have continuous variable curvature \cite{wang2024soft}. The use of a growing spine incorporates the ability to grow and retract with milli-particle granular jamming to create a new type of mechanism that allows self-controllable variable curvature capability. This newly designed soft continuum robot presents a new modeling problem to be solved. 

\begin{figure*}[t]
 \centering
 \includegraphics[width=\textwidth]{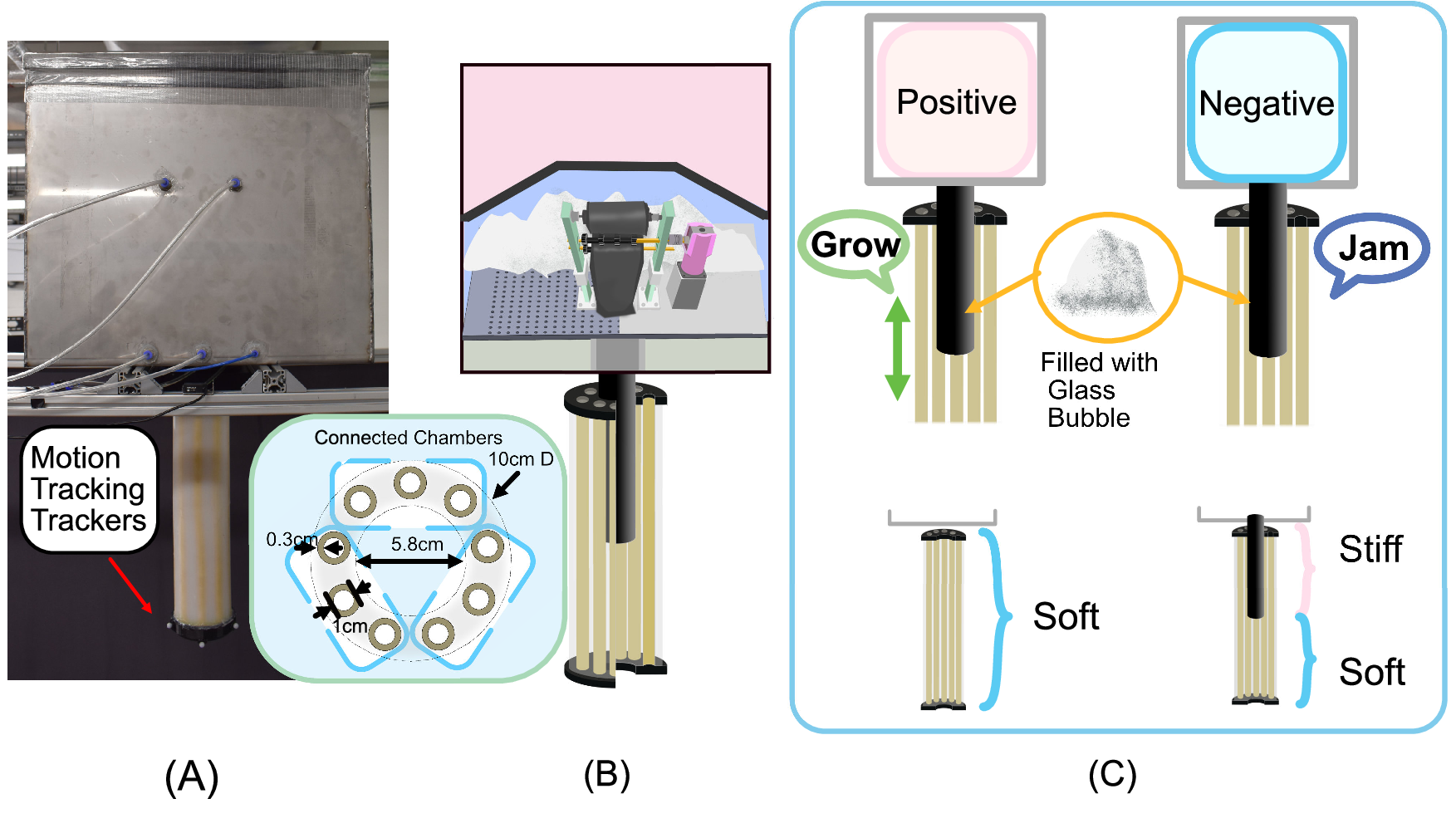} 
 \caption{Detailed system design. (A) Soft continuum robot system with motion tracking markers on the end-effector. (B) Cross-section area to show the inner structure of the robot. Inside the steel box, airtight fabric divides the volume into two. The bottom volume contains glass bubbles and a length control mechanism. The filter paper on top of the pegboard prevents glass bubbles from going into the vacuum chamber when jamming. (C) The state transitions to allow the robot to reconfigure. The growing spine is filled with granules. It changes length inside the continuum robot and stiffness when jammed using negative pressure.}
 \label{fig:2}
\end{figure*}

The constant curvature model is a common approach to model the shape of continuum robots. However, due to its assumption that the backbone of the continuum robot is represented as a constant curvature shape, this approach ignores the external load effect. Piecewise constant curvature model divides the continuum into multiple sections and each section is assumed to have constant curvature \cite{della2020model}\cite{webster2010design}.  More approaches utilizing variable curvature modeling methods has been explored, adapted for specific continuum robot designs like multi-back bones and planar soft continuum robots \cite{chen2021variable}\cite{huang2021kinematic}. In terms of variable curvature modeling, the Cosserat rod model is popular due to its high accuracy and account for external loads \cite{till2019real}\cite{wang2022data}. By dividing the continuum robot section into many small parts and applying force and moment balances along the small part, the Cosserat rod model has been proven to be effective on soft continuum robots \cite{renda2018discrete}\cite{gilbert2019validation}\cite{li2023discrete}.

For the silicone-based soft continuum robot with fiber-reinforced chambers, a principle of virtual work combined with Cosserat rod model has also been experimented for both linear and non-linear silicone elasticity \cite{trivedi2008geometrically}\cite{sadati2017geometry}\cite{shiva2019elasticity}. A similar method is also implemented for a Miniature Eversion Growing Robot for control \cite{wu2023vision}\cite{sadati2021tmtdyn}. In this paper, we adapt the Cosserat rod model specifically for our designed soft continuum robot with a growing spine \cite{wang2024soft}. Experiments have been performed on its growing spine for measuring stiffness value at different lengths. Under the different configurations of our soft continuum robot, we have tracked end-effector positions to make comparison with our Cosserat rod model for validation.

\section{Soft Continuum Robot Design}

The soft continuum robot has two main components as shown in Figure \ref{fig:2}(A). The sealed chamber rests above the aluminum frame with a soft continuum robot attached underneath it. The soft continuum robot has a length of 40cm long. It has a hollow inner part that allows the spine to grow and retract. 

In Figure \ref{fig:2}(B), the cross-section area of the sealed chamber has two volumes divided by a flexible airtight TPU-coated fabric. This volume is also connected with the growing robot made of airtight TPU-coated fabric \cite{hawkes2017soft}. We have a customized length-control unit for accurate position control of the growing spine. The bottom volume also contains glass bubble material to allow granular jamming \cite{bakarich2022pump}. Using air-tight fabric to partition the volume in a sealed chamber offers the advantage of achieving a tight fit of granules during granular jamming. 

By varying the growing spine length, the stiffness distribution across the soft continuum robot is changed as shown in Figure \ref{fig:2}(c). This approach enables a seamless adjustment of the stiffening portion, facilitating a continuous variable curvature capability.

\section{Math-modelling Based on Cosserat Rod Theory}

\subsection{Cosserat Rod Model Approach}

Along the center-line of the rod, we define several variables for the position $p(s)$, orientation $R(s)$, internal force $n(s)$, internal moment $m(s)$, linear strain rate $v(s)$, and angular strain rate $u(s)$ of the continuum robot. When considering any point $s$ along the rod, we can apply force and moment balance equations to an infinitesimal section. The explicit cosserat staic rod model is presented as following:
 \begin{align}
     \begin{split}
          \dot{p} & = Rv\\
          \dot{R} & = R\widehat{u}\\
          \dot{n} & = -f\\
          \dot{m} & = -\dot{p}\times{n}-l
      \end{split}
      \label{eq:1}
 \end{align}

The $\hat{}$ symbol represent a mapping of $R^3$ to $SO(3)$. The $dot{p}$ is the change rate of the position $p(s)$ with respect to $s$. Similarly, the $\dot{R}$, $\dot{n}$, and $\dot{m}$ represent the change rate of rotation, internal force, and internal moment with respect to $s$. 

We have used linear constitution equations to map the internal force n(s) and moment m(s) of a rod with linear strain rate $v(s)$ and angular strain rate $u(s)$ as:
 \begin{align}
     \begin{split}
          v = K_{se}^{-1}R^Tn + v^{*}\\
          u = K_{bt}^{-1}R^Tm + u^{*}
      \end{split}
      \label{eq:2}
 \end{align}

\begin{table}[t]
\centering
\caption{Physical Parameters of Soft continuum Robot}
\label{table1}
\begin{tabular}{lc}
\toprule
Parameter & Value \\
\midrule
Outer Radius $r_o$ & 5 cm \\
Inner Radius $r_i$ & 2.9 cm \\
Chamber Radius $r_c$ & 0.5 cm \\
Chamber Path Radius $r_{path}$ & 4cm\\
Density $\rho$ & \(1300 \, \text{kg/m}^3\) \\
Young's Modulus $E$ & 0.507147 MPa\\
Shear Modulus $G$ & $E/3$ MPa\\
Length $L$ & 40 cm\\
External Load $F_{external}$ & 53 gram\\
\bottomrule
\end{tabular}
\end{table}

where $K_{se}$ represent the stiffness matrix for extension and shear, $K_{bt}$ represent the stiffness matrix for bending and torsion. The configurations used as reference are denoted by $v^{*}$ and $u^{*}$. The soft continuum is fixed to the aluminum frame on the one end, so the $u^{*}$ is [0;0;0] and $v^{*}$ is [0;0;1] to represent strain along the continuum robot. Specifically, $K_{se}$ and $K_{bt}$ are related to the material property of the continuum robot:
\begin{align}
    \begin{split}
        K_{se}(s) & =diag(GA(s), GA(s), EA(s))\\
        K_{bt}(s) & =diag(EI_{xx}(s), EI_{yy}(s)), EI_{zz}(s))\\
    \end{split}
    \label{eq:3}
\end{align}

The cross-sectional area of the rod is denoted by $A(s)$ and the shear modulus by $G(s)$. The Young's modulus is represented by $E(s)$. The second moments of area in the corresponding axis are given by $I_{xx}$ and $I_{yy}$, where $I_{zz}$ equals $I_{xx}$ plus $I_{yy}$. The detailed physical parameters of the soft continuum robot are listed in Table \ref{table1}.

\subsection{Soft Continuum Robot Pneumatic Actuation}
 The soft continuum robot is actuated through pneumatic pressure. The elongation of the chambers makes the continuum robot extend and bend. The actuation force on the continuum robot is formulated as following:
 \begin{align}
    \begin{split}
        n_{P} & = \sum_{i} P_i A_{effect} R e_3\\
        m_{P} & = \sum_{i} \hat{path_i}P_i A_{effect} R e_3  \\
    \end{split}
    \label{eq:4}
\end{align}
There are 9 chambers inside the continuum robot as shown in Figure \ref{fig:2}(A). The chamber location at the local frame is fixed across the length of the soft continuum robot. $path_i$ represents the coordinate information for the $i$th chamber. $R$ represents the orientation of the pressure. $P_i$ represents the pressure value for each chamber.

\subsection{Boundary Condition}
The initial condition for our soft continuum robot is $p(0)$ = [0;0;0], and $R(0)$ is 3 by 3 identity matrix. The force and moment caused by the pneumatic pressure and external force are at the length L (free end) on the continuum robot. 

 \begin{align}
    \begin{split}
        n(L) & = n_P + F_{external}\\
        m(L) & = m_P + L_{external} \\
    \end{split}
    \label{eq:5}
\end{align}

The $F_{external}$ and $L_{external}$ represent the external force and moment on the tip of the soft continuum robot.

\subsection{Solving Boundary Value Problem}

The spatial integration of the Cosserat rod along its length uses Euler's integration method, which divides the rod into $N$ number of points. We take the position of the rod at point $i$ on the rod as an example:
\begin{align}
    \begin{split}
    p_{i+1}=p_i+ds*p_{s_i},
    \end{split}
    \label{eq:6}
\end{align}
where $p_{i+1}$ is the position of the next point on the rod and $p_{s_i}$ is obtained from equation  \eqref{eq:1}.

To solve this boundary value problem, we apply the shooting method to reduce the residual terms for given pressure $P_i$ in each chamber. We can represent the initial guess and residual term as following:
\begin{align}
    \begin{split}
        Guess &= [n(0) \ m(0)\ ]\\
        Error &= [(E^F)^T \ (E^M)^T]\\
     \end{split}
     \label{eq:7}
\end{align}

\section{Model Variable Curvature Soft Continuum Robot}
\subsection{Modelling of the Growing Spine}

\begin{figure}[t!]
 \centering
 \includegraphics[width=0.9\linewidth]{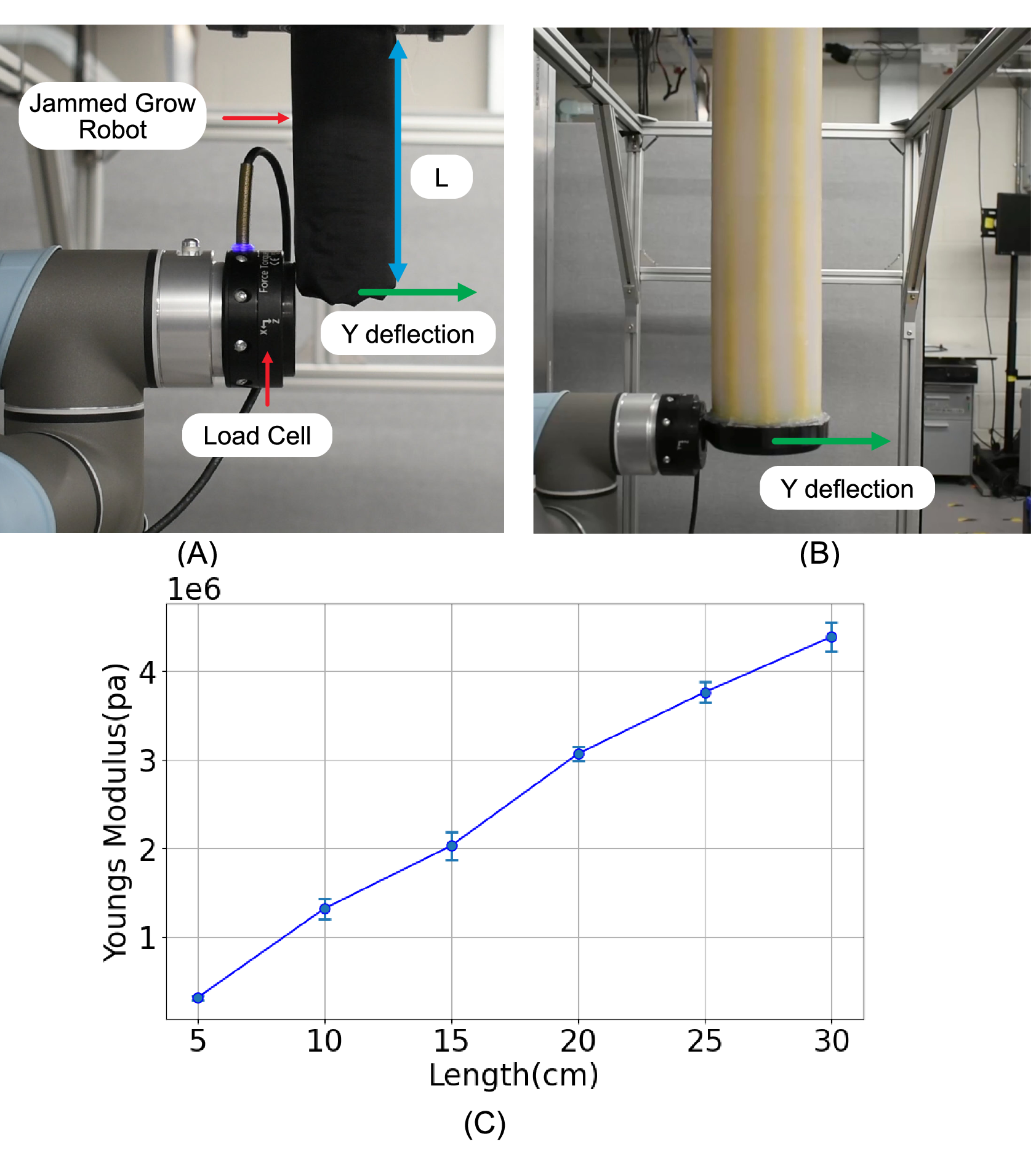} 
 \caption{(A) Experiment setup for testing the stiffness of the growing robot using UR5 robotic arm and ROBOTiq ft300-s. (B) Experiment setup for testing the stiffness of the soft continuum robot}
 \label{fig:3}
\end{figure}

To determine the jammed growing spine properties, we have performed experiments shown in Figure \ref{fig:3}. By mounting a ROBOTiq ft300-s load cell on Universal Robots UR5, we can precisely control the robot's movement and record the corresponding force value from the load cell. To gain a better understanding of the continuously stiffness distribution properties of the growing spine, we confidently evaluate its capabilities at length of 5cm, 10cm, 15cm, 20cm, 25cm, and 30cm with small intervals in between.

We have employed the Euler-Bernoulli beam theory for modelling the jammed-growing spine. Given a small displacement $y$, when applied force at position $x$ along the beam, we calculate Young's modulus as:
\begin{equation}
y = \frac{F \times (3L-x) x^2}{6 \times E \times I}
\label{eq:8}
\end{equation}

The second moment of area for the jammed growing spine has circular shape, $I = \frac{\pi r^4}{4}$, substituting $I$ in equation \ref{eq:8}, the Young's modulus $E$ with $x = L$ can be obtained as:
\begin{equation}
E = \frac{4 \times F \times L^3}{3 \times \pi r^4 \times y }
\label{eq:9}
\end{equation}

similarly, the Young's modulus for the silicone-based continuum robot can also be obtained from the equation \eqref{eq:8} by changing to soft continuum robot second moment of area. This value is presented in Table \ref{table1}.

As shown in Figure \ref{fig:3}(c), the Youngs' modulus of the growing spine is presented in the figure. 

\begin{table}[t]
\centering
\caption{Young's Modulus Value for Jammed Growing Spine}
\label{table2}
\begin{tabular}{cccc}
\hline
Length (cm) & 5 & 10 & 15 \\
E (MPa) & $0.318$  & $1.323$ & $2.032$\\
\hline
Length (cm) & 20 & 25 & 30 \\
E (MPa) & $3.069$ & $3.763$  & $4.389$ \\
\hline
\end{tabular}
\end{table}

 We can find clear trend that as the growing robot grow longer, the Young's modulus value also significantly increases. Due to the soft continuum robot's large dimension, the gravity increases the overall stiffness of the jammed growing spine as well. 

\subsection{Comprehensive Modelling with Combined stiffness}

When the growing spine has a 0 cm length inside the soft continuum robot, the mathematical modeling is identical to the model mentioned in the previous section. The situation has changed when the growing spine starts to grow inside the hollow soft continuum robot.

\begin{figure}[t!]
 \centering
 \includegraphics[width=0.9\linewidth]{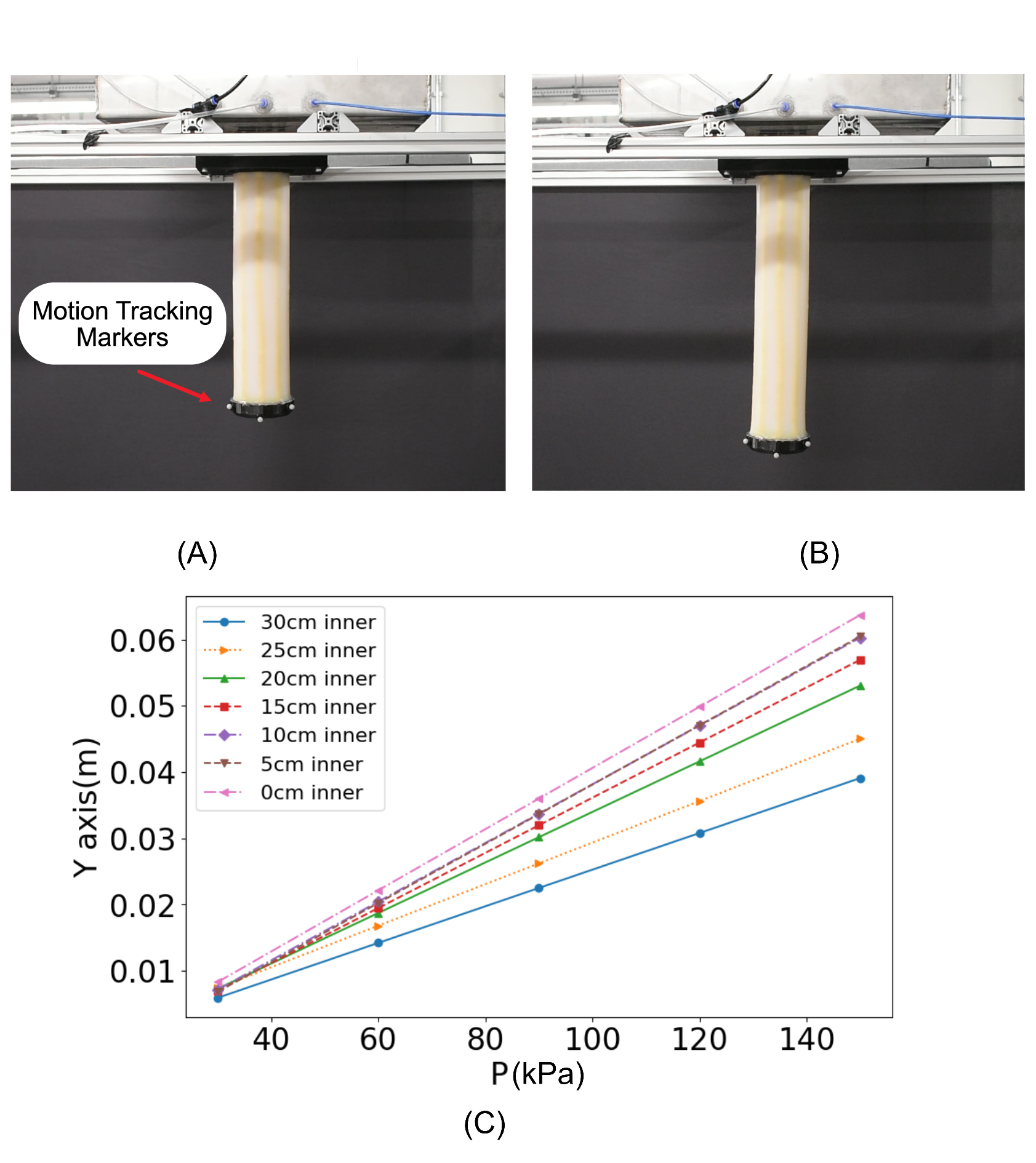} 
 \caption{(A) Experiment setup for testing the extending of the soft continuum robot with different growing robot configuration inside. The end-effector position is captured through the motion tracking system. (B) An elongated soft continuum robot. (C) The relationship between the elongation of the robot and applied pressure for different growing spine configurations inside the continuum robot}
 \label{fig:4}
\end{figure}

Through our experiments exploring the extension capabilities of the growing spine within the soft hollow continuum robot at varying lengths, we have found that the overall extension ability of the soft continuum robot is decreased from Figure \ref{fig:4}(c). During this experiment, the continuum robot was subjected to pressurization across all nine chambers, resulting in a uniform elongation along its axis. The pressure ranged from 30kPa to 150kPa, increasing in increments of 30kPa. Due to the high friction coefficient of the silicone rubber, when the growing robot is in place of the soft continuum robot, it creates a large amount of friction force to prevent it from extending. 

While this relationship is extremely complicated, we make an assumption in the modeling approach that the portion of the growing spine is combined with the soft continuum robot with no sliding between them. We can then formalize the combined stiffness using the following equation:

\begin{equation}
E_{\text{eq}} = \frac{V_c}{V_{\text{total}}} E_c + \frac{V_s}{V_{\text{total}}} E_s
\label{eq:10}
\end{equation}

\begin{figure*}[t!]
 \centering
 \includegraphics[width=0.9\linewidth]{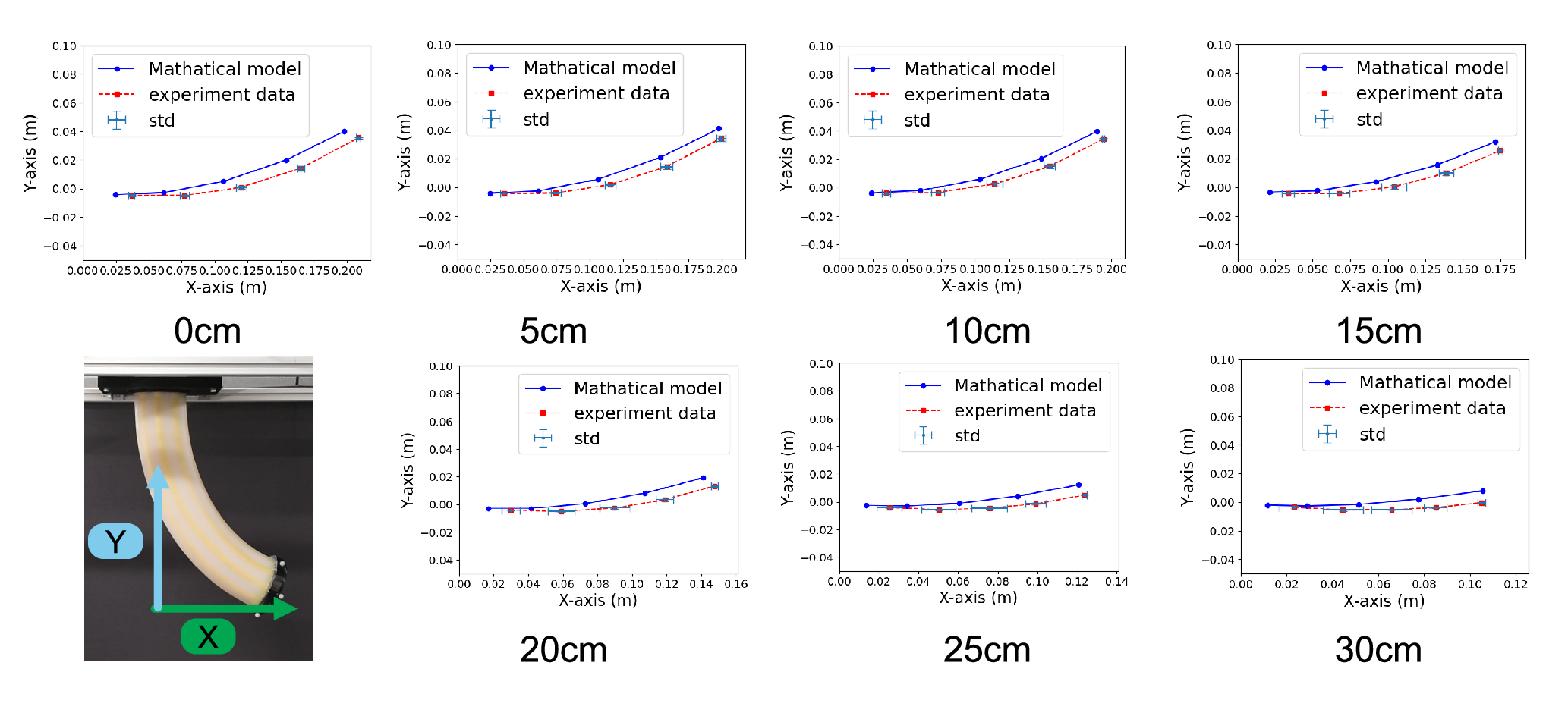} 
 \caption{Comparison between the mathematical model and experimental results covers the range of soft continuum robot configurations from 0 to 30 cm, and pressurized from 50 to 250 kPa with increments of 50 kPa.}
 \label{fig:5}
\end{figure*}

The combined stiffness is calculated based on the Young's modulus of the jammed growing spine in Table 
\ref{table2} and the soft continuum robot. Since the length and cross-section area are fixed. This volume ratio is fixed for the combined stiffness portion.

To incorporate this updated stiffness into the mathematical model, we modified the $K_{se}$ and $K_{bt}$ values in equations \eqref{eq:2} and \eqref{eq:3} since it is directly related to the combined Young's Modulus and strain rate change. 

When using Euler's method to integrate along the continuum robot, if the current position is less than the growing spine length, we used the combined stiffness from equation \eqref{eq:10} as the jammed growing robot significantly increased the stiffness at the portion. Otherwise, we use the silicone Young's modulus in Table \ref{table1}. By treating the continuum robot as a single-section continuum robot, we reduce the step for modifying the boundary conditions and simplify the modeling approach.

\subsection{Model Calibration}
We assume there is no slide between the growing spine surface and the soft continuum robot's inner surface in the ideal situation. We obtained the stiffness information from real experiments. To reduce the modeling error, we have selected $A_{effect}$ as the only parameter we calibrate for the model, as seen in the previous paper \cite{gilbert2019validation}. 

This value is initially based on the geometry area of pressure chamber $A_{norm}$. After calibration, for the soft continuum robot with 0 cm growing spine length, $A_{effect} = 1.5 A_{norm}$. As the growing spine grows inside the continuum robot. This coefficient becomes 1.5, 1.7, 1.9, 2, 2.15, 2.4 for growing spine length of 5cm, 10cm, 15cm, 20cm, 25cm and 30cm.

\begin{table}[t]
\centering
\caption{Mathematical modelling error}
\label{table3}
\begin{tabular}{ccccc}
\hline
Length (cm) & 0 & 5 & 10 & 15\\
Error in X axis (m) & 0.0127 & 0.0078 & 0.0090 & 0.0096\\
Error in Y axis (m) & 0.0034 & 0.0037 & 0.0032 & 0.0037\\
\hline
Length (cm) & 20 & 25 & 30 &\\
Error in X axis (m) & 0.0132 & 0.011 &
 0.00999 & \\
Error in Y axis (m) & 0.0035 & 0.0042 &
 0.0044 & \\
\hline
\end{tabular}
\end{table}

\section{Experiments for Validation Math Model}

\subsection{Experimental Setup}

Six motion tracking cameras from OptiTrack have been installed in the room, operating at a tracking frequency of 100Hz. The tracking error is minimal with an average of less than 0.5 mm in each axis. To create the bending motion of the soft continuum robot, we have pressurized 3 individual chambers shown in Figure \ref{fig:2}(A), leaving 6 individual chambers unpressurized.
The soft-continuum robot's end-effector position is tracked at various pressurized levels, including 50kPa, 100kPa, 150kPa, 200kPa, and 250kPa.

\subsection{Experimental Results And Discussion}

The soft continuum robot end effector exhibits varying trajectories due to the different growing spine configurations. As depicted in Figure \ref{fig:5}, it is evident that the length of the growing spine inside the soft continuum robot significantly reduces the soft continuum robot end-effector position in the X-axis direction. The differences in the resulting end-effector positions can be visualized in Figure \ref{fig:1} as well.

The variable curvature of this type of soft continuum robot presents a significant challenge for the mathematical model. We extensively validate the our model through a series of bending experiments performed on the soft continuum robot, with various configurations ranging from 0cm to 30cm growing spine in increments of 5cm. The control inputs are range from 50kPa to 250kPa with increments of 50kPa.  

The results show that the comprehensive model with combined stiffness is able to match the experimental results very well. The maximum average error in the direction of X-axis 1.32 cm. The maximum average error in the direction of Y axis is  0.42 cm as shown in Table \ref{table3}. Compared to the size of soft continuum robot of 40cm long, the percentage error is 3.3\% in X axis and 1.05\% in the direction of Y-axis. Due to the bending of the robot toward X-axis, the error in X-axis is larger than then error in Y-axis.

Based on the experimental data, we have a high level of confidence in the Cosserat rod model, as it demonstrates a strong correlation with the observed results. Meanwhile, from the elongation experiments in Figure~\ref{fig:4}. We could also conclude that the soft continuum robot exhibits linear elasticity when experiencing relatively small deformation. We calibrated the $A_{effect}$ in our math model. By employing this approach, we can leverage the unmodeled friction that occurs as the soft continuum robot slides against the growing spine.

\section{Conclusion}
We have modeled a soft continuum robot with self-controllable variable curvature using an adapted Cosserat rod model with combined stiffness method. Our approach has been thoroughly validated through extensive experiments that involved testing different configurations of the soft continuum robot and input pressure levels. The results show high accuracy of this modeling approach. The novelty comes from incorporating the combined stiffness during the implementation of Euler's method. This way the problem of modeling different stiffness portions of the continuum robot can be solved under the same boundary condition of a single-section continuum robot thus reducing the problem's complexity. Future work will focus on exploring dynamic modeling based on Cosserat rod variable curvature methods and control strategies for the variable curvature soft continuum robot.

\addtolength{\textheight}{-12cm}   








\bibliography{references}

\begin{thebibliography}{10}

\bibitem{guan2023trimmed}
Q.~Guan, F.~Stella, C.~Della~Santina, J.~Leng, and J.~Hughes, ``Trimmed helicoids: an architectured soft structure yielding soft robots with high precision, large workspace, and compliant interactions,'' {\em npj Robotics}, vol.~1, no.~1, p.~4, 2023.

\bibitem{fras2015new}
J.~Fra{\'s}, J.~Czarnowski, M.~Macia{\'s}, J.~G{\l}{\'o}wka, M.~Cianchetti, and A.~Menciassi, ``New stiff-flop module construction idea for improved actuation and sensing,'' in {\em 2015 IEEE international conference on robotics and automation (ICRA)}, pp.~2901--2906, IEEE, 2015.

\bibitem{haggerty2023control}
D.~A. Haggerty, M.~J. Banks, E.~Kamenar, A.~B. Cao, P.~C. Curtis, I.~Mezi{\'c}, and E.~W. Hawkes, ``Control of soft robots with inertial dynamics,'' {\em Science Robotics}, vol.~8, no.~81, p.~eadd6864, 2023.

\bibitem{rao2023modeling}
P.~Rao, C.~Pogue, Q.~Peyron, E.~Diller, and J.~Burgner-Kahrs, ``Modeling and analysis of tendon-driven continuum robots for rod-based locking,'' {\em IEEE Robotics and Automation Letters}, 2023.

\bibitem{ma2023inspired}
K.~Ma, X.~Chen, J.~Zhang, Z.~Xie, J.~Wu, and J.~Zhang, ``Inspired by physical intelligence of an elephant trunk: Biomimetic soft robot with pre-programmable localized stiffness,'' {\em IEEE Robotics and Automation Letters}, vol.~8, no.~5, pp.~2898--2905, 2023.

\bibitem{zhang2023preprogrammable}
J.~Zhang, Y.~Li, Z.~Kan, Q.~Yuan, H.~Rajabi, Z.~Wu, H.~Peng, and J.~Wu, ``A preprogrammable continuum robot inspired by elephant trunk for dexterous manipulation,'' {\em Soft Robotics}, vol.~10, no.~3, pp.~636--646, 2023.

\bibitem{yang2020geometric}
C.~Yang, S.~Geng, I.~Walker, D.~T. Branson, J.~Liu, J.~S. Dai, and R.~Kang, ``Geometric constraint-based modeling and analysis of a novel continuum robot with shape memory alloy initiated variable stiffness,'' {\em The International Journal of Robotics Research}, vol.~39, no.~14, pp.~1620--1634, 2020.

\bibitem{pogue2022multiple}
C.~Pogue, P.~Rao, Q.~Peyron, J.~Kim, J.~Burgner-Kahrs, and E.~Diller, ``Multiple curvatures in a tendon-driven continuum robot using a novel magnetic locking mechanism,'' in {\em 2022 IEEE/RSJ International Conference on Intelligent Robots and Systems (IROS)}, pp.~472--479, IEEE, 2022.

\bibitem{stella2023prescribing}
F.~Stella, J.~Hughes, D.~Rus, and C.~Della~Santina, ``Prescribing cartesian stiffness of soft robots by co-optimization of shape and segment-level stiffness,'' {\em Soft Robotics}, 2023.

\bibitem{cianchetti2014soft}
M.~Cianchetti, T.~Ranzani, G.~Gerboni, T.~Nanayakkara, K.~Althoefer, P.~Dasgupta, and A.~Menciassi, ``Soft robotics technologies to address shortcomings in today's minimally invasive surgery: the stiff-flop approach,'' {\em Soft robotics}, vol.~1, no.~2, pp.~122--131, 2014.

\bibitem{arleo2023variable}
L.~Arleo, L.~Lorenzon, and M.~Cianchetti, ``Variable stiffness linear actuator based on differential drive fiber jamming,'' {\em IEEE Transactions on Robotics}, 2023.

\bibitem{clark2022malleable}
A.~B. Clark and N.~Rojas, ``Malleable robots: Reconfigurable robotic arms with continuum links of variable stiffness,'' {\em IEEE Transactions on Robotics}, vol.~38, no.~6, pp.~3832--3849, 2022.

\bibitem{kang2023soft}
J.~Kang, S.~Lee, and Y.-L. Park, ``Soft bending actuator with fiber-jamming variable stiffness and fiber-optic proprioception,'' {\em IEEE Robotics and Automation Letters}, 2023.

\bibitem{wang2024soft}
X.~Wang, Q.~Lu, D.~Lee, Z.~Gan, and N.~Rojas, ``A soft continuum robot with self-controllable variable curvature,'' {\em IEEE Robotics and Automation Letters}, 2024.

\bibitem{della2020model}
C.~Della~Santina, R.~K. Katzschmann, A.~Bicchi, and D.~Rus, ``Model-based dynamic feedback control of a planar soft robot: trajectory tracking and interaction with the environment,'' {\em The International Journal of Robotics Research}, vol.~39, no.~4, pp.~490--513, 2020.

\bibitem{webster2010design}
R.~J. Webster~III and B.~A. Jones, ``Design and kinematic modeling of constant curvature continuum robots: A review,'' {\em The International Journal of Robotics Research}, vol.~29, no.~13, pp.~1661--1683, 2010.

\bibitem{chen2021variable}
Y.~Chen, B.~Wu, J.~Jin, and K.~Xu, ``A variable curvature model for multi-backbone continuum robots to account for inter-segment coupling and external disturbance,'' {\em IEEE Robotics and Automation Letters}, vol.~6, no.~2, pp.~1590--1597, 2021.

\bibitem{huang2021kinematic}
X.~Huang, J.~Zou, and G.~Gu, ``Kinematic modeling and control of variable curvature soft continuum robots,'' {\em IEEE/ASME Transactions on Mechatronics}, vol.~26, no.~6, pp.~3175--3185, 2021.

\bibitem{till2019real}
J.~Till, V.~Aloi, and C.~Rucker, ``Real-time dynamics of soft and continuum robots based on cosserat rod models,'' {\em The International Journal of Robotics Research}, vol.~38, no.~6, pp.~723--746, 2019.

\bibitem{wang2022data}
X.~Wang and N.~Rojas, ``A data-efficient model-based learning framework for the closed-loop control of continuum robots,'' in {\em 2022 IEEE 5th International Conference on Soft Robotics (RoboSoft)}, pp.~247--254, IEEE, 2022.

\bibitem{renda2018discrete}
F.~Renda, F.~Boyer, J.~Dias, and L.~Seneviratne, ``Discrete cosserat approach for multisection soft manipulator dynamics,'' {\em IEEE Transactions on Robotics}, vol.~34, no.~6, pp.~1518--1533, 2018.

\bibitem{gilbert2019validation}
H.~B. Gilbert and I.~S. Godage, ``Validation of an extensible rod model for soft continuum manipulators,'' in {\em 2019 2nd IEEE International Conference on Soft Robotics (RoboSoft)}, pp.~711--716, IEEE, 2019.

\bibitem{li2023discrete}
H.~Li, L.~Xun, G.~Zheng, and F.~Renda, ``Discrete cosserat static model-based control of soft manipulator,'' {\em IEEE Robotics and Automation Letters}, vol.~8, no.~3, pp.~1739--1746, 2023.

\bibitem{trivedi2008geometrically}
D.~Trivedi, A.~Lotfi, and C.~D. Rahn, ``Geometrically exact models for soft robotic manipulators,'' {\em IEEE Transactions on Robotics}, vol.~24, no.~4, pp.~773--780, 2008.

\bibitem{sadati2017geometry}
S.~H. Sadati, S.~E. Naghibi, A.~Shiva, Y.~Noh, A.~Gupta, I.~D. Walker, K.~Althoefer, and T.~Nanayakkara, ``A geometry deformation model for braided continuum manipulators,'' {\em Frontiers in Robotics and AI}, vol.~4, p.~22, 2017.

\bibitem{shiva2019elasticity}
A.~Shiva, S.~H. Sadati, Y.~Noh, J.~Fra{\'s}, A.~Ataka, H.~W{\"u}rdemann, H.~Hauser, I.~D. Walker, T.~Nanayakkara, and K.~Althoefer, ``Elasticity versus hyperelasticity considerations in quasistatic modeling of a soft finger-like robotic appendage for real-time position and force estimation,'' {\em Soft Robotics}, vol.~6, no.~2, pp.~228--249, 2019.

\bibitem{wu2023vision}
Z.~Wu, S.~H. Sadati, K.~Rhode, and C.~Bergeles, ``Vision-based autonomous steering of a miniature eversion growing robot,'' {\em IEEE Robotics and Automation Letters}, 2023.

\bibitem{sadati2021tmtdyn}
S.~H. Sadati, S.~E. Naghibi, A.~Shiva, B.~Michael, L.~Renson, M.~Howard, C.~D. Rucker, K.~Althoefer, T.~Nanayakkara, S.~Zschaler, {\em et~al.}, ``Tmtdyn: A matlab package for modeling and control of hybrid rigid--continuum robots based on discretized lumped systems and reduced-order models,'' {\em The International Journal of Robotics Research}, vol.~40, no.~1, pp.~296--347, 2021.

\bibitem{hawkes2017soft}
E.~W. Hawkes, L.~H. Blumenschein, J.~D. Greer, and A.~M. Okamura, ``A soft robot that navigates its environment through growth,'' {\em Science Robotics}, vol.~2, no.~8, p.~eaan3028, 2017.

\bibitem{bakarich2022pump}
S.~E. Bakarich, R.~Miller, R.~A. Mrozek, M.~R. O'Neill, G.~A. Slipher, and R.~F. Shepherd, ``Pump up the jam: Granular media as a quasi-hydraulic fluid for independent control over isometric and isotonic actuation,'' {\em Advanced Science}, vol.~9, no.~15, p.~2104402, 2022.

\end{thebibliography}
\bibliographystyle{ieeetr}

\end{document}